\PassOptionsToPackage{hyphens}{url}
\documentclass[11pt, logo, onecolumn, copyright]{nvidiatechreport}
\renewcommand{\today}{2026-09-09}

\usepackage[round]{natbib}
\usepackage{newunicodechar}
\newunicodechar{—}{\textemdash}
\usepackage{listings}
\usepackage{breakcites}
\usepackage{float}
\usepackage{afterpage}
\setlist[itemize]{topsep=1pt,itemsep=1pt}
\setlist[enumerate]{topsep=1pt,itemsep=1pt}
\usepackage{xparse}
\RenewDocumentCommand{\paragraph}{s o m}{%
  \par\medskip\noindent\textbf{#3}\quad\ignorespaces
}

\title{An Open Recipe for IMO Gold: Training Nemotron for Olympiad Mathematics}
\author{\large Ivan Moshkov, Stephen Ge, George Armstrong, Wei Du, Sadegh Mahdavi, Igor Gitman}
\date{}

\begin{document}

\begin{abstract}
\large \textbf{Abstract.}
We study how model post-training and test-time inference design affect natural-language proof generation for hard olympiad mathematics. Starting from Nemotron 3 Ultra, we train two specialist checkpoints using supervised fine-tuning and reinforcement learning, and evaluate checkpoint choice, verification, and refinement. Based on these findings, we present an open-model test-time-compute pipeline. The system operates entirely in natural language, with no formal prover, external tools, or internet access. Three Nemotron 3 Ultra checkpoints - the general-availability model and two post-trained specialists - power an iterative search that generates, verifies, and refines candidate proofs; a separate high-compute stage then selects each final submission. The system scored 30 out of 42 points at IMO 2026, reaching the gold-medal threshold. We release the two post-trained checkpoints as well as the training data, the training and inference code, the submitted solutions, and Nemotron-IMO-Bench, a new benchmark of 200 novel olympiad-level problems.

\end{abstract}

\maketitle

\section{Introduction}
\label{sec:introduction}

The International Mathematical Olympiad (IMO) has historically been one of the most celebrated intellectual competitions in the world, and over the past few years it has also become a grand test of mathematical problem-solving ability for AI systems.\footnote{See the \href{https://imo-grand-challenge.github.io/}{IMO Grand Challenge} and \href{https://aimoprize.com/}{AIMO Prize} initiatives.} AI systems combining generative methods with formal verification reached silver-medal level in 2024 \citep{hubertOlympiadlevelFormalMathematical2026,trinhSolvingOlympiadGeometry2024}, and in 2025 natural-language models reached gold \citep{luong2025geminiimo,openaiWeAchievedGold2025}. This report studies how model post-training and test-time inference choices affect natural-language proof generation, and describes the resulting Nemotron 3 Ultra \citep{nvidiaNemotron3Ultra2026}-based system submitted to the 2026 competition.

We combine Nemotron 3 Ultra checkpoints trained for proof generation, verification, grading, and refinement with a high-compute inference strategy. Our system operates entirely in natural language: it uses no formal prover, external tool, or internet access. It receives the competition organizers' LaTeX problem statements and produces natural-language solutions for submission. We evaluate the effects of checkpoint choice, verification, and multi-model inference on a 30-problem development set.

A central goal of this work is reproducibility and future extensibility. We release as much of the system as possible: post-trained checkpoints, training data, training and inference code, submitted solutions, and detailed compute resource accounting. Our contributions are:

\begin{itemize}
\item An empirical study of model post-training and test-time inference choices, including checkpoint performance, verification, and the full ensemble.
\item An open release of two post-trained checkpoints, training data, training and inference code, and submitted solutions.
\item Nemotron-IMO-Bench, comprising 200 novel olympiad-level problems and the 30-problem development set used in this report.
\item Resource accounting for the competition run, including compute hours and generated-token counts.
\end{itemize}

\section{Release Artifacts}
\label{sec:releases}

All artifacts are gathered in the Hugging Face collection
\href{https://huggingface.co/collections/nvidia/nemotron-labs-imo-2026}{\nolinkurl{nvidia/nemotron-labs-imo-2026}},
together with the base Nemotron-3-Ultra-GA model.

\paragraph{Checkpoints.} We release Nemotron-3-Ultra-SFT as
\href{https://huggingface.co/nvidia/Nemotron-3-Labs-Ultra-Math-SFT}{\nolinkurl{nvidia/Nemotron-3-Labs-Ultra-Math-SFT}}
and Nemotron-3-Ultra-RL as
\href{https://huggingface.co/nvidia/Nemotron-3-Labs-Ultra-Math-RL}{\nolinkurl{nvidia/Nemotron-3-Labs-Ultra-Math-RL}},
both under the OpenMDW-1.1 license of the base model.

\paragraph{Training data.} The SFT corpus (Section~\ref{sec:training-sft}) is
released as
\href{https://huggingface.co/datasets/nvidia/Nemotron-Math-Proofs-v3-SFT}{\nolinkurl{nvidia/Nemotron-Math-Proofs-v3-SFT}}
and the RL problem set (Section~\ref{sec:training-rl}) as
\href{https://huggingface.co/datasets/nvidia/Nemotron-Math-Proofs-v3-RL}{\nolinkurl{nvidia/Nemotron-Math-Proofs-v3-RL}},
both under the CC BY 4.0 license.

\paragraph{Benchmark.} We release \textbf{Nemotron-IMO-Bench}, a collection of
200 novel olympiad-level problems created in collaboration with Professor Titu
Andreescu, as
\href{https://huggingface.co/datasets/nvidia/Nemotron-IMO-Bench}{\nolinkurl{nvidia/Nemotron-IMO-Bench}}
under the CC BY 4.0 license.

\paragraph{Code and submitted proofs.} The inference pipeline, the script
assembling the 30-problem development set, and the proofs submitted to IMO 2026
are available in NeMo-Skills at
\url{https://github.com/NVIDIA-NeMo/Skills/tree/main/recipes/nemotron-imo-tts}.
The RL training recipe is available in NeMo-RL at
\url{https://github.com/NVIDIA-NeMo/RL/blob/imo-26-ultra-v3/docs/guides/nemotron-3-ultra-imo.md}.
The SFT stage follows the same training pipeline as Nemotron-3-Ultra-GA,
described in the Nemotron 3 Ultra technical
report~\citep{nvidiaNemotron3Ultra2026}.

\section{Related Work}
\label{sec:related-work}

\subsection{Neuro-symbolic and formal methods for math olympiads} The first competitive results at the IMO for AI came from systems based on neuro-symbolic solving and formal methods. AlphaGeometry \citep{trinhSolvingOlympiadGeometry2024} used a neural language model to guide a symbolic engine. At IMO 2024, AlphaProof and AlphaGeometry 2 \citep{hubertOlympiadlevelFormalMathematical2026} combined to score one point below the human gold level cutoff. Additional Lean-based provers such as DeepSeekProver, GoedelProver, and SeedProver \citep{renDeepSeekProverV2AdvancingFormal2025,linGoedelProverV2ScalingFormal2025,chenSeedProverDeepBroad2025} advanced the open-weight frontier of neural formal theorem proving, progressively improving on canonical benchmarks including miniF2F and PutnamBench \citep{zhengMiniF2FCrosssystemBenchmark2022,tsoukalasPutnamBenchEvaluatingNeural2024}. The scalability of these systems depended heavily on reliable symbolic verification, either through Lean or domain-specific geometry tools, to generate large training corpora and search over highly branching spaces in challenging problems.

\subsection{Natural-language breakthroughs for proofs} IMO 2025 brought major advances in natural-language proving. Gemini Deep Think \citep{luong2025geminiimo} and an experimental OpenAI model \citep{openaiWeAchievedGold2025} both reached the gold-medal level with end-to-end natural-language systems, producing full-credit solutions to each of the five problems they solved. The results demonstrated not only the proof-generation abilities of frontier LLMs, but also the importance of verification and final-solution selection \citep{mahdaviBrainsVsBytes2025,guoRightNotEnough2025a}.

\subsection{Feedback-driven refinement loops and math agents}
\citet{huangWinningGoldIMO2025} demonstrated that a model-agnostic verification-and-refinement pipeline based on models publicly available at the time could also achieve the gold-medal level. Related approaches in domains outside mathematical proof include Self-Refine \citep{madaanSelfRefineIterativeRefinement2023} and RLEF \citep{gehringRLEFGroundingCode2025} in code execution. Nomos \citep{jinNomos2025} achieved top results on Putnam 2025 with a post-trained model and a reasoning harness with parallel generation, scoring, and consolidation without feedback-based refinement.

\subsection{Proof-search systems and test-time scaling}
DeepSeekMath-V2 \citep{shaoDeepSeekMathV2SelfVerifiableMathematical2025} trained generator, verifier, and meta-verifier models and scaled verification compute as part of a high-compute search setup. Aletheia \citep{fengAutonomousMathematicsResearch2026}, a math research agent powered by Gemini Deep Think with explicit Generator, Verifier, Reviser subagents in an iterative harness, pushed the frontier from Olympiads to research-level mathematics. Nemotron-Cascade 2 \citep{yangNemotronCascade2PostTraining2026} demonstrated that compact models can also approach the capabilities of frontier open models in the proof generation domain.

\section{Models and Training}
\label{sec:models-training}

\subsection{Models}
\label{sec:models}

Our system uses three Nemotron-3-Ultra 550B-A55B checkpoints as the inference pipeline workers (Section \ref{sec:pipeline}). The \textbf{Nemotron-3-Ultra-GA} is the general-availability checkpoint, used unchanged. Starting from it, we post-train two specialists: \textbf{Nemotron-3-Ultra-SFT} via supervised fine-tuning (Section \ref{sec:training-sft}) and \textbf{Nemotron-3-Ultra-RL} via reinforcement learning (Section \ref{sec:training-rl}). The pipeline uses these checkpoints in three roles: generation (proposing candidate proofs), verification (judging a candidate's correctness and producing feedback), and refinement (revising a candidate using that feedback). The exact role assignment is given in Section \ref{sec:pipeline}; this section describes the checkpoints and their training. Both post-trained checkpoints are part of the open release (Section \ref{sec:releases}).

\subsection{Training}
\label{sec:training}
\subsubsection{Supervised Fine-Tuning}
\label{sec:training-sft}

We perform a long-context supervised fine-tuning stage starting from the
\texttt{Nemotron-3-Ultra-GA} checkpoint%
\footnote{\url{https://huggingface.co/nvidia/NVIDIA-Nemotron-3-Ultra-550B-A55B-BF16}}.
The model is fine-tuned on a proof-focused SFT corpus with a maximum sequence
length of 425{,}984 tokens. We optimize the standard per-token cross-entropy
loss in BF16 precision.

\paragraph{Proof data generation.}
We construct the proof-focused corpus with a multi-stage synthetic-data
pipeline designed to supervise both long-form proof construction and proof
verification. We begin with 15{,}879 challenging mathematical proof problems
from the AoPS subset of \path{Nemotron-Math-Proofs-v1}\footnote{\url{https://huggingface.co/datasets/nvidia/Nemotron-Math-Proofs-v1}}, selected using prior pass-rate
evaluations to concentrate generation on difficult problems. For each problem,
we use \texttt{DeepSeek-V4-Pro}%
~\citep{xu2026deepseek} in Max inference mode to generate multiple initial proof
attempts (see Appendix~\ref{app:generation-prompts}), with a maximum generation
length of 400K tokens. Problems not judged
to be fully solved are passed through up to three additional refinement rounds.
Each refinement conditions on earlier attempts and verifier feedback, and asks
the model to identify gaps, repair invalid reasoning, and produce a revised
solution (see Appendix~\ref{app:refinement-prompt}). Refinement trajectories
use a maximum length of 350K tokens.

In parallel, verifier trajectories assess candidate proofs and assign a final
score in $\{0, 0.5, 1\}$ (see Appendix~\ref{app:verification-prompt}), while
meta-verifier trajectories assess the reliability of self-evaluations and
verifier judgments (see Appendix~\ref{app:meta-verification-prompt}). We remove incomplete,
malformed, empty, and length-capped generations, as well as examples with
invalid response structure. Proof and refinement examples are retained only
when they contain non-empty visible \texttt{Solution} and \texttt{Self
Evaluation} sections; verification examples must contain a parseable final
score and evaluate the visible proof. To avoid a training distribution dominated
by incorrect proofs, we retain all valid score-$0.5$ and score-$1$ verifier
traces and deterministically subsample score-$0$ traces. The resulting corpus contains 414{,}890 quality-filtered examples over 15{,}818 unique problems: 58{,}543 proof-generation traces, 67{,}971 refinement traces, 236{,}360 verification traces, and 52{,}016 meta-verification traces. The mixture
therefore teaches the model to construct proofs, diagnose logical gaps, revise
unsuccessful approaches, and assess proof validity.

The SFT run uses 512 GB200 GPUs, with tensor parallelism 8, context parallelism 32,
expert parallelism 64, expert tensor parallelism 1, and pipeline parallelism 1.
The global batch size is 64 and the micro-batch size is 1, corresponding to
approximately 1.6K optimizer steps for one pass over the packed data.

We use AdamW with $\beta_1 = 0.9$, $\beta_2 = 0.95$, weight decay 0.1, and
gradient clipping at 1.0. The learning rate warms up for 1{,}024 samples,
approximately 1\% of the training set, to a peak value of $1.5 \times 10^{-5}$,
followed by cosine decay to $2 \times 10^{-6}$ over the rest of training. To
support the 426K-token context length, we enable selective recomputation for
MoE layers and fine-grained activation offloading for MoE activations.

We select the checkpoint at step 1300 from this SFT stage based on evaluation
performance on 133 proof-based problems drawn from IMO-ProofBench%
~\citep{luongRobustMathematicalReasoning2025} and recent math competitions.

\subsubsection{Reinforcement Learning}
\label{sec:training-rl}
Starting from the Nemotron-3-Ultra-GA checkpoint, we train the model using reinforcement learning (RL) to improve its proof generation abilities.

\textbf{Data.} For proof-generation RL, we draw problems from Nemotron-Math-Proofs-v1.\footnote{\href{https://huggingface.co/datasets/nvidia/Nemotron-Math-Proofs-v1}{Nemotron-Math-Proofs-v1 on Hugging Face}.} We retain those that Nemotron-3-Ultra solves in one to three of four attempts, as judged by DeepSeek-V3.2-Speciale. This yields 9,597 problems for training the proof generator.

\textbf{RL Rewards.} We largely follow the reward design of DeepSeekMath-V2~\citep{shaoDeepSeekMathV2SelfVerifiableMathematical2025}. However, we remove the self-analysis reward by setting $\alpha=1$ and $\beta=0$.

\textbf{RL Algorithm.} We use an asynchronous reinforcement learning (RL) framework built on NeMo-RL~\citep{nemo-rl}\footnote{See the \href{https://github.com/NVIDIA-NeMo/RL/blob/imo-26-ultra-v3/docs/guides/nemotron-3-ultra-imo.md}{NeMo RL code and IMO 2026 Ultra training recipe}.} To improve training efficiency, we adopt an algorithm similar to PipeLineRL~\citep{piche2025pipelinerl}. On the inference side, we maintain a fixed pool of prompts in flight at all times. Completed sequences are continuously passed to the training engine whenever a full training batch becomes available. We further apply dynamic sampling~\citep{yu2026dapo} to prevent the effective batch size on the trainer side from decreasing due to samples with zero advantage.
We use truncated importance sampling as in Nemotron-3-Ultra~\citep{nvidiaNemotron3Ultra2026}. To control the entropy, we mask out low-probability tokens in positive samples when entropy exceeds $0.4$.

\textbf{Hyperparameters.} Each training batch contains 128 prompts, with 16 trajectories sampled per prompt, resulting in a global training batch size of 2,048 trajectories. We use AdamW with $\beta_1=\beta_2=0.90$, a learning rate of $3\times10^{-6}$, and no weight decay. The maximum trajectory age is four optimizer steps, beyond which the older datapoints are discarded. Each training run uses 128 trainer nodes, 128 inference nodes, and 16 judge nodes, with up to 160 prompt groups concurrently in flight and a maximum sequence length of 131,072 tokens. Each node contains four NVIDIA GB200 GPUs.

\begin{figure}[H]
\centering
\includegraphics[width=\linewidth]{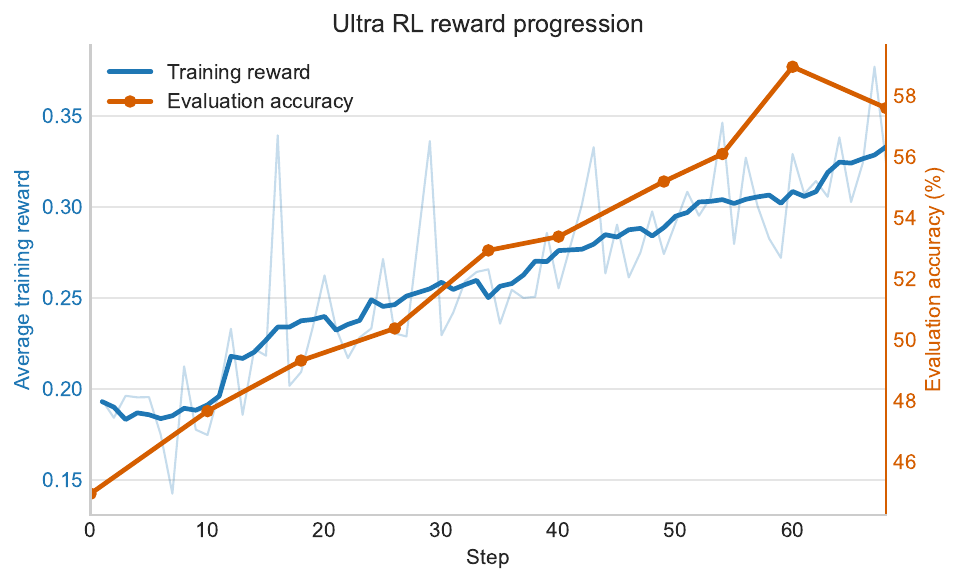}
\caption{\textbf{RL reward progression.}
Reinforcement learning yields rapid improvements in both the training reward and performance on the evaluation set, as judged by GPT-5.5.}
\label{fig:rl-rewards}
\end{figure}

\textbf{Evaluation.} We evaluate the model on 133 proof-based problems drawn from IMO-ProofBench~\citep{luongRobustMathematicalReasoning2025}, the 2025 IMO and Putnam competitions, and recent national and international mathematical olympiads held in 2025. We use GPT-5.5 with xhigh reasoning effort as the evaluator. Figure~\ref{fig:rl-rewards} shows the training reward and evaluation performance over the course of RL.

\section{Submitted Pipeline}
\label{sec:pipeline}

Our IMO submission system consists of two stages. The first stage is a high-compute search (Section~\ref{sec:high-compute-search}): an ensemble of three Nemotron-3-Ultra checkpoints proposes candidate proofs, a checkpoint panel scores every candidate and produces natural-language critiques, and subsequent rounds revise the most promising candidates using those critiques. All candidates, together with their scores and critiques, accumulate in a per-problem proof pool, and the search for a problem ends at the close of the round in which a candidate is unanimously accepted by the verification panel, or after a fixed round budget.
The second stage (Section~\ref{sec:multi-model-verification}) re-evaluates the resulting finalists with a substantially larger judgment budget and selects the proof submitted to the competition. Section~\ref{sec:imo-results} reports the official result and accounts for the compute consumed.

\subsection{High-Compute Search}
\label{sec:high-compute-search}

We use an iterative generate-verify-refine procedure inspired by
DeepSeekMath-V2~\citep{shaoDeepSeekMathV2SelfVerifiableMathematical2025}.
Proofs are stored with their verifier scores and feedback in a per-problem
proof pool. If no proof is accepted, the next round refines candidates from
this pool. The process runs independently for each problem for at most eight
rounds.

The submitted system is a multi-model ensemble. Generation uses
Nemotron-3-Ultra-GA, Nemotron-3-Ultra-RL, and Nemotron-3-Ultra-SFT. Round 1 produces 384 proof
attempts: each checkpoint samples 16 attempts from each of eight complementary
generation prompts. The templates instruct the model to follow different solution strategies (lemma-first decomposition, route comparison, counterexample search, etc.; see Appendix~\ref{app:generation-prompts}). Multiple prompts are used to decorrelate round-1 generations and diversify
the attempted approaches. We split the round-1 budget across checkpoints
instead of spending it on more attempts from one of them. In our experiments a
second checkpoint solves problems the first cannot, whereas doubling the
attempts of a single checkpoint adds little (Section~\ref{sec:ensemble-results}).

Search-time verification uses Nemotron-3-Ultra-RL and Nemotron-3-Ultra-SFT. The verifier is
reference-free and assigns a score of 1 to a complete and correct proof, 0.5
to a generally correct proof with minor errors or omissions, and 0 to a proof
with fatal errors or severe omissions. Each checkpoint produces eight
independent judgments for every proof, giving 16 equally weighted judgments.
A judgment is valid if it terminates with a parseable final score. A proof is accepted only when a complete panel of 16 valid judgments is available and every judgment assigns a score of 1. Here and throughout the report, \emph{accepted} means that a proof satisfies this internal verification criterion; it does not by itself imply correctness under independent or official grading. The exact verifier prompt is given in
Appendix~\ref{app:verification-prompt}.

If no proof is accepted, the system selects up to 16 highest-ranked proofs from
the global proof pool and constructs one refinement prompt for each selected
proof, together with up to eight verifier critiques. Every prompt is sent to
all three generation checkpoints, with four outputs sampled from each, for 192
refinement attempts per round. Refined proofs are verified and added to the
same pool. The exact refinement prompt is given in
Appendix~\ref{app:refinement-prompt}.

\subsection{Final Candidate Selection}
\label{sec:multi-model-verification}

During search, verification is designed to support iterative improvement:
the verifier assigns correctness scores and produces actionable critiques that
guide subsequent refinement rounds. The search applies early stopping at the
checkpoint level: once a generation checkpoint produces an accepted proof, no
further candidates are sampled from it, while generation from the remaining
checkpoints continues until the end of the current round. A problem therefore
ends the search with up to three finalists, one per generation checkpoint. If
no proof is accepted within the round budget, the highest-ranked proof in the
pool is the sole finalist.

Final candidate selection has a different objective: ranking the remaining
proofs for submission. We evaluate every finalist with Nemotron-3-Ultra-GA,
Nemotron-3-Ultra-RL, and Nemotron-3-Ultra-SFT using a reference-free IMO-style
judge prompt. Whereas the search-time verifier prompt is designed to guide
refinement, the final-selection prompt is designed for IMO-style scoring.
The prompt adapts the proof-evaluation methodology described by
\citet{dekoninckmatharena2026} to a reference-free setting: it instructs the
judge to identify the milestones required for a complete solution and assign
an integer score from 0 to 7 according to what the submitted proof establishes.
The full prompt is provided in Appendix~\ref{app:imo-judge-prompt}.

For every finalist, each checkpoint produces 16 independent IMO-style
judgments, yielding 48 judgments per finalist. Finalists are ranked by the
mean of these 48 scores, with ties broken in favor of the shorter proof text.
The top-ranked proof is selected for submission.

\subsection{IMO 2026 Results}
\label{sec:imo-results}

\paragraph{Official result.} The system participated officially in IMO 2026 and scored 30 out of 42 points, above the gold-medal cutoff of 29; all submitted proofs were graded by official IMO graders. The submissions received full credit on Problems 1, 2, 4, and 5, and one point each on Problems 3 and 6. Table~\ref{tab:resources} breaks the run down by problem. For each problem, it lists the points awarded and the search round in which the submitted proof was accepted. It also reports the cumulative tokens and GB200 GPU-hours consumed, measured at two moments: when the submitted proof became available, and when computation on the problem stopped. All six submitted proofs were found within approximately 707M generated tokens and 1,464 GPU-hours. Completing the rounds already in flight brought the full competition run to approximately 2.31B tokens and 4,800 GPU-hours.

\begin{figure}[t]
\centering
\includegraphics[width=\linewidth]{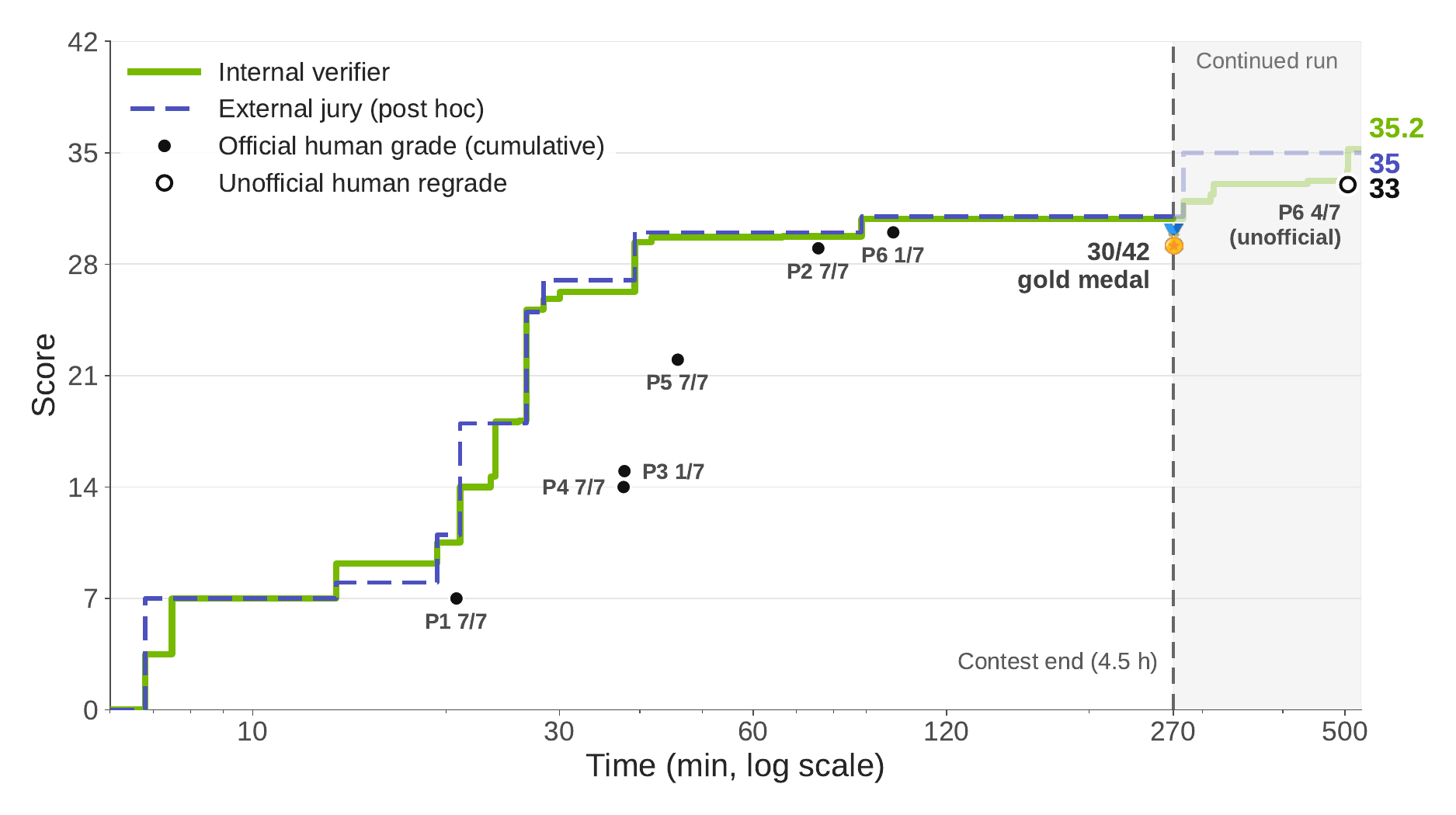}
\caption{\textbf{Competition score over time (log scale).} The contest spans two 4.5-hour sessions (P1--P3 on day 1, P4--P6 on day 2); each problem is plotted against elapsed time since the start of its session. Green: the summed per-problem internal verifier score, scaled to the 42-point range. Blue dashed: post-hoc scores from the independent model jury (Section~\ref{sec:experimental-proof-scoring}) on proofs that advanced the internal frontier; jury compute is excluded from the time axis. Filled circles: cumulative official score, plotted when each submitted proof cleared the final-selection panel (Section~\ref{sec:multi-model-verification}). The vertical line marks the 4.5-hour contest cutoff, when the submission stood at 30/42 (gold cutoff: 29). The shaded region shows continued search after the cutoff; the later P6 proof received 4/7 in an unofficial independent human regrade.}
\label{fig:imo-run}
\end{figure}

\paragraph{Competition run.} Figure~\ref{fig:imo-run} traces the run. On each contest day, the three problem searches ran concurrently and shared the full competition GPU allocation; the figure plots each problem against elapsed time since the start of its session. A proof counts as available only once its full verification panel has completed. All submitted proofs were finalized early in the run: the four full-credit proofs passed the final-selection panel of Section~\ref{sec:multi-model-verification} within the first 76 minutes, and the remaining two within 100 minutes. The search plateaued thereafter, before the 4.5-hour competition deadline. The solid curve shows the internal estimate available during the run: for each problem, the best mean search-verifier score in its pool, summed over problems and scaled to the 42-point range.

The dashed curve was not part of the competition system. It was computed after the run, as part of our post-hoc analysis. The independent model jury (Section~\ref{sec:experimental-proof-scoring}) graded every proof that had advanced the internal frontier, with the grading time excluded from the time axis. The two signals track each other closely throughout the run, suggesting that the search-time verifier provides a reliable real-time proxy for independent evaluation. At the contest cutoff, however, both verifiers assigned a score of roughly 32 points, two points above the official result of 30. This discrepancy is entirely attributable to Problems 3 and 6, where both model-based evaluations credited proofs that received only one point from the official graders. The error therefore appears to reflect a shared blind spot in model-based verification rather than noise specific to the search-time verifier.

\paragraph{Continued run.} The shaded region of Figure~\ref{fig:imo-run} shows the same system running past the contest cutoff. In round 8, after 8 h 25 m of total search time, Problem 6 search produced a new solution. The internal verifier did not accept it, but it scored higher than the contest submission. This required an additional 610M tokens and 890 GPU-hours beyond the contest window (Table~\ref{tab:resources}). As part of the post-hoc analysis, we asked a panel of independent human mathematicians to grade this solution. The graders awarded it 4 out of 7 points. The grading was performed without access to the official marking schemes and is not an official IMO result. Under this assessment, the total would rise to 33 points. Problem 3 did not improve.

\begin{table}[t]
\centering
\begin{tabular}{lrrrrrr}
\toprule
& & Proof & \multicolumn{2}{c}{Tokens} & \multicolumn{2}{c}{GPU-hours} \\
\cmidrule(lr){4-5}\cmidrule(lr){6-7}
Problem & Points & round & to proof & run total & to proof & run total \\
\midrule
P1 & 7 & R1 & 5.57M & 218M & 11.7 & 459.2 \\
P2 & 7 & R2 & 227M  & 318M & 479.3 & 670.2 \\
P3 & 1 & R1 & 106M  & 641M & 240.9 & 1,387.7 \\
P4 & 7 & R1 & 20.4M & 231M & 42.9 & 487.4 \\
P5 & 7 & R1 & 36.4M & 253M & 76.7 & 533.9 \\
P6 & 1 & R2 & 312M  & 650M & 612.9 & 1,246.2 \\
\midrule
Competition total & 30 & -- & 707M & 2.31B & 1,464.4 & 4,784.6 \\
\midrule
P6 continued$^\dagger$ & 4 (unofficial) & R8
    & 1.26B & 1.37B
    & 2,135.7 & 2,296.3 \\
\bottomrule
\end{tabular}
\caption{Per-problem resource accounting for the competition run on GB200
GPUs. Token counts cover generation, refinement, and verification, rounded
to three significant figures. ``To proof'': cumulative consumption when the
submitted proof became available. ``Run total'': consumption when
computation stopped---the end of the round in which the proof was accepted
for P1, P2, P4, and P5, and the 4.5-hour contest cutoff for P3 and P6.
$^\dagger$Found after 8\,h\,25\,m, past the contest cutoff; the 4/7 grade
is an unofficial regrade by a panel of independent human graders without access to
the official marking schemes.}
\label{tab:resources}
\end{table}
\section{Experimental Setup}
\label{sec:setup}

The remainder of the report evaluates the pipeline and the design decisions behind it. Because rerunning the full submitted system is expensive, most ablations use single-checkpoint configurations on a 30-problem development set. This section defines the shared experimental infrastructure: the evaluation data (Section \ref{sec:evaluation-dataset}), the single-model search configuration used by the ablations (Section \ref{sec:single-model-search}), and the independent jury used for scoring (Section \ref{sec:experimental-proof-scoring}). Section~\ref{sec:experiments} reports the experiments.

\subsection{Evaluation Dataset}
\label{sec:evaluation-dataset}

Working with Professor Titu Andreescu, a mathematics educator and olympiad
problem author, we created
\textbf{Nemotron-IMO-Bench}, a collection of 200 novel olympiad-level
problems. The problems were written for this benchmark and have not been published before, so they do not appear in the training data of the models we evaluate. Evaluating the full benchmark with our high-compute pipeline was too
expensive for rapid iteration. We therefore formed a 30-problem development
set containing 20 problems from Nemotron-IMO-Bench and 10 problems from recent
competitions.

We used a preliminary Nemotron-3-Ultra-GA run to estimate difficulty and selected
problems spanning the full observed range. The development set contains 3
easy, 7 medium, 10 hard, and 10 unsolved problems.
Easy problems have a substantial first-round acceptance rate; medium problems have
only one or two accepted proofs in the first round or were first accepted in
round 2; hard problems were first accepted between rounds 3 and 8; and
unsolved problems have no accepted proof. The set contains
6 algebra, 8 combinatorics, 8 geometry, and 8 number-theory problems.

We release Nemotron-IMO-Bench under the CC BY 4.0 license; release details are summarized in
Section~\ref{sec:releases}. The 10 development-set problems drawn from recent competitions are publicly available on Hugging Face; we provide a script that assembles the full 30-problem development set from the released benchmark and these public sources.

\subsection{Single-Model Baselines}
\label{sec:single-model-search}

For single-model baselines, one checkpoint performs generation, refinement,
and search-time verification. Round 1 samples 128 proof attempts. Each unique proof receives 64 verification attempts from the same checkpoint. Attempts that fail to return a parseable score are discarded. A proof is accepted only if at least 60 attempts remain and all of them assign a score of 1. Verification stops early after accumulating eight judgments with a score below 1.

Each later round constructs 32 refinement prompts from the global top-32 proof
pool and samples four outputs per prompt, producing 128 refinement attempts.
As in the submitted pipeline, search runs for at most eight rounds and returns
the highest-ranked proof in the final pool if no proof satisfies the acceptance
criterion.

\subsection{Proof Scoring}
\label{sec:experimental-proof-scoring}

The ablation studies use a jury of GPT-5.5, Gemini 3.1 Pro, and Claude Opus
4.8, separate from the Nemotron panel used during the competition. Each model
independently scores the proof from 0 to 7 using the same IMO-style judge
prompt as the final-selection panel in
Section~\ref{sec:multi-model-verification} (Appendix~\ref{app:imo-judge-prompt}).

We follow MathArena's jury procedure~\citep{dekoninckmatharena2026}. If the
initial scores differ by at most two points, the minimum score is used.
Otherwise, every judge grades the proof again after seeing the first-round
assessments, and the minimum second-round score is used. Incomplete panels are
rerun. Unlike MathArena, our evaluation is reference-free and does not use
normalization, tool calls, or official-solution-derived rubrics. The exact
reconciliation wrapper is given in Appendix~\ref{app:reconciliation-prompt}.

We report two quantities. The per-round cumulative score sums the independent-jury scores of problems with an internally accepted proof by that round; problems without an accepted proof contribute zero. Parentheses report the cumulative number of internally accepted problems. After the final round, a problem still without an accepted proof proceeds with the highest-ranked proof in its final pool, as in the submitted pipeline. We score this proof and report the total over all 30 problems. In the experimental results, we call a problem \emph{solved} when the independent jury assigns its selected proof full credit.

\section{Experiments}
\label{sec:experiments}

We ablate three components separately from the submitted system:
single-checkpoint proof search, checkpoint verification, and the complete
ensemble. Unless stated otherwise, all
experiments use the 30-problem development set, the eight-round search budget,
and the scoring protocol from Section~\ref{sec:setup}.

\subsection{Single-Checkpoint Pipeline Performance}
\label{sec:checkpoint-performance}

We independently run Nemotron-3-Ultra-GA, Nemotron-3-Ultra-RL, and Nemotron-3-Ultra-SFT using
the single-model baseline from Section~\ref{sec:single-model-search}. Each
checkpoint performs generation, verification, and refinement, so this
experiment compares complete single-checkpoint pipelines rather than generation
quality in isolation. Table~\ref{tab:checkpoint-performance} reports the
cumulative independent-jury score after each search round.

\begin{table}[H]
    \centering
    \small
    \begin{tabular}{lrrrr}
        \toprule
        Round & Nemotron-3-Ultra-GA & Nemotron-3-Ultra-RL & Nemotron-3-Ultra-SFT & Ensemble \\
        \midrule
        R1 &  34 (5)  &  47 (7)  &  70 (10) &  91 (13) \\
        R2 &  72 (12) &  82 (12) &  98 (14) & 160 (23) \\
        R3 &  93 (15) & 109 (16) & 112 (16) & 167 (24) \\
        R4 & 114 (18) & 137 (20) & 126 (18) & 167 (24) \\
        R5 & 128 (20) & 137 (20) & 140 (20) & 167 (24) \\
        R6 & 142 (22) & 151 (22) & 147 (21) & 167 (25) \\
        R7 & 142 (22) & 151 (22) & 147 (21) & 167 (25) \\
        R8 & 145 (23) & 152 (23) & 147 (21) & 167 (25) \\
        R8 + fallback & 162 (23) & 180 (23) & 165 (21) & 188 (25) \\
        \bottomrule
    \end{tabular}
    \caption{Cumulative independent-jury score on the 30-problem development set. Parentheses give the cumulative number of problems with an internally accepted proof. A problem begins contributing in the round in which its proof is first accepted. The final row reports the end-to-end result: if no proof is accepted after eight rounds, the highest-ranked proof remaining in the pool is used as the finalist, following the same rule as in the submitted pipeline.}
    \label{tab:checkpoint-performance}
\end{table}

Both post-trained checkpoints outperform Nemotron-3-Ultra-GA.
Nemotron-3-Ultra-SFT performs best in the first round, while
Nemotron-3-Ultra-RL achieves the best overall single-checkpoint result. The
R1--R8 rows score only problems for which the search verifier accepts a proof.
If no proof is accepted after eight rounds, the pipeline still returns the
highest-ranked proof from the final pool as the sole finalist. The last row
also scores these finalists, giving the end-to-end result over all 30
problems.

\subsection{Checkpoint Verification Performance}
\label{sec:verifier-eval}

The search-time verifier (Section~\ref{sec:high-compute-search}) uses Nemotron-3-Ultra-RL and Nemotron-3-Ultra-SFT, eight judgments each, and accepts a proof only if all 16 judgments assign score~1. To test these choices separately from generation, we audit all three checkpoints as verifiers on 300 proofs from the development-set ensemble run (Section~\ref{sec:ensemble-results}): the 25 proofs accepted by the submitted panel and 275 non-accepted proofs stratified by search-verifier score. Each proof receives eight judgments from each checkpoint under the prompt of Appendix~\ref{app:verification-prompt}; the RL and SFT judgments come from the search itself, and the GA judgments were collected for this analysis. As ground truth, we score every proof with the independent jury of Section~\ref{sec:experimental-proof-scoring} and count a proof as correct when it receives 7 points. Because the set is enriched for difficult, high-scoring proofs, the rates compare rules on the same proofs; they are not prevalence estimates for the full pool.
 
\begin{table}[t]
\centering
\small
\begin{tabular}{@{}llrr@{}}
\toprule
Panel & Rule & False accept (\%) & False reject (\%) \\
\midrule
\multicolumn{4}{@{}l}{\emph{Single checkpoint}} \\
Nemotron-3-Ultra-GA  & 8/8 & 31.6 [17.9, 45.4] & 22.8 [11.0, 35.5] \\
Nemotron-3-Ultra-RL  & 8/8 & 12.4 [4.7, 20.4]  & 64.2 [53.8, 72.5] \\
Nemotron-3-Ultra-SFT & 8/8 & 4.5 [0.6, 9.0]    & 69.1 [62.1, 75.3] \\
\midrule
\multicolumn{4}{@{}l}{\emph{Unanimous mixed panels}} \\
GA+SFT                      & 16/16          & 3.4 [0.0, 7.3]          & 75.6 [67.6, 81.7] \\
\textbf{RL+SFT (submitted)} & \textbf{16/16} & \textbf{1.1 [0.0, 2.7]} & \textbf{81.3 [75.8, 85.0]} \\
GA+RL+SFT                   & 24/24          & 1.1 [0.0, 2.7]          & 82.9 [77.5, 86.7] \\
\midrule
\multicolumn{4}{@{}l}{\emph{Submitted panel, relaxed threshold}} \\
RL+SFT & 14/16 & 17.0 [6.5, 27.9]  & 59.4 [48.4, 69.0] \\
RL+SFT & 12/16 & 25.4 [10.5, 42.3] & 30.9 [20.0, 42.2] \\
\bottomrule
\end{tabular}
\caption{Verifier operating points on the 300-proof audit set. A rule $k/n$ accepts a proof when at least $k$ of $n$ judgments assign score~1. False accept is the share of jury-incorrect proofs a rule accepts and false reject the share of jury-correct proofs it rejects, with jury score~7 as correct; brackets are 95\% intervals from 2{,}000 problem-clustered bootstrap resamples.}
\label{tab:verifier-panels}
\end{table}
 
The single-checkpoint rows of Table~\ref{tab:verifier-panels} show that the three checkpoints differ mainly in selectivity. Nemotron-3-Ultra-GA accepts half of the audit set, including roughly a third of the jury-incorrect proofs; Nemotron-3-Ultra-SFT accepts the fewest proofs and is the most precise; Nemotron-3-Ultra-RL falls between. For a verifier that decides when the search stops, the balance between the two error types matters more than accuracy, because they are not equally costly. A false accept ends the search for a problem on an invalid proof, and no later stage generates new candidates. A false reject only delays: the proof keeps its high mean score, typically remains among the top-ranked refinement candidates, and stays eligible as the fallback submission if nothing is accepted (Section~\ref{sec:multi-model-verification}). The search-time verifier therefore uses the two selective checkpoints and demands unanimity.
 
The remaining rows show that no tested panel improves both rates and that there is little room to relax either choice. Lowering the threshold to 14/16 raises the false-accept rate from about 1\% to 17\%, so that in this set each correct proof the relaxed rule recovers is matched by an incorrect one it admits. Adding GA's eight judgments to the panel does not help, since a permissive verifier rarely vetoes what RL and SFT have both accepted; the only two proofs it removes are correct. What drives the panel's precision is that the two selective checkpoints catch each other's errors. SFT alone accepts eight incorrect proofs; RL's eight judgments reject six of them, taking the false-accept rate from 4.5\% to 1.1\%, whereas GA's eight judgments reject only two (3.4\%). Under unanimity every judgment is another chance to veto, and RL's vetoes fall where SFT's do not.
 
Of the 25 accepted proofs, 23 receive jury score~7. One receives~6 for a minor quantifier formality. One receives~0: its argument rests on an invalid column-permutation symmetry step, and an explicit counterexample refutes the claimed optimum. Every checkpoint gave both proofs eight full-score judgments, so no unanimity rule over these 24 judgments would have caught them, whereas GA alone would have rejected two correct accepts. Within this panel, then, no choice of rule separates the two errors from the 23 correct accepts.

\subsection{Full Ensemble Pipeline}
\label{sec:ensemble-results}

Finally, we evaluate the submitted ensemble from Section~\ref{sec:pipeline}
on the development set. The ensemble reaches its final accepted-only score
within three rounds and outperforms every single-checkpoint pipeline.

When we also score the highest-ranked final-pool proof for each problem without
an accepted proof, the ensemble finishes eight points ahead of
Nemotron-3-Ultra-RL. The ensemble run is not compute-matched to the
single-checkpoint runs, so Table~\ref{tab:attempt-scaling} isolates the design
question behind it, whether the round-1 budget is better spent on more attempts
from one checkpoint or on attempts from other checkpoints.

\begin{table}[t]
\centering
\small
\begin{tabular}{@{}lrrrr@{}}
\toprule
& & Accepted & \multicolumn{2}{c}{Jury score} \\
\cmidrule(lr){4-5}
Round-1 pool & Tokens (B) & problems & Accepted only & All 30 \\
\midrule
RL 128                    & 0.76 & 13 &  87 & 135 \\
RL 256                    & 1.59 & 14 &  92 & 139 \\
SFT 64                    & 0.80 &  7 &  43 & 121 \\
SFT 128                   & 1.60 & 11 &  68 & 127 \\
\midrule
RL 128 + SFT 64           & 1.55 & 15 &  98 & 159 \\
RL 128 + SFT 128          & 2.36 & 18 & 116 & 159 \\
RL 128 + GA 128           & 1.06 & 13 &  87 & 140 \\
RL 128 + SFT 128 + GA 128 & 2.67 & 18 & 116 & 158 \\
\bottomrule
\end{tabular}
\caption{Round-1 attempt scaling on the development set. Each row names the
checkpoints in the pool and the number of attempts per problem from each. A
proof is accepted when all eight Nemotron-3-Ultra-RL verifier judgments assign
score~1, and proofs are ranked by mean RL judgment with ties broken by length.
``Accepted only'' sums the independent-jury scores of the accepted
proofs; ``All 30'' scores the top-ranked proof of every problem. Tokens count
proof generation only.}
\label{tab:attempt-scaling}
\end{table}

Every pool in Table~\ref{tab:attempt-scaling} is judged with the RL checkpoint
alone. This rule is more permissive than the deployed 2-model judge
(Table~\ref{tab:verifier-panels}), so we compare pools by the jury score of
their top-ranked proofs and report both the accepted-only and the
all-problem score as well as solved problem count. RL~128 is the ensemble's round-1 RL share under the eight
generation prompts, and RL~256 adds the round-1 attempts of the RL
single-checkpoint baseline (Section~\ref{sec:single-model-search}). The SFT and
GA rows are the ensemble's shares under the same eight RL verifier seeds;
SFT~64 keeps 8 of the 16 attempts of each prompt.

Doubling the RL attempts barely moves the pool. A second set of 128 attempts
yields one more accepted problem and a few more jury points. Spending the same
tokens on 64 SFT attempts is worth considerably more, because the SFT
checkpoint reaches problems that RL does not. Of the 18 problems the combined
RL~128 + SFT~128 pool accepts, six are accepted from both checkpoints, seven
only from RL and five only from SFT, and doubling the RL attempts to 256
recovers just one of the five.

GA adds no problem that RL or SFT do not already accept in round-1. Pooling its attempts
with RL and SFT shifts the all-problem score by a point, since a larger pool can
change which proof ranks first. However, we still use GA for all rounds to allow it's attempts enter the candidate pool for later rounds, where it is a useful source of refinement diversity.

\subsection{Approaches That Did Not Improve the Final System}

We also explored several alternatives to proof selection and refinement, including cross-proof context, alternative proof-ranking rules, refinement from multiple diverse proofs, and triage-based routing. These variants sometimes accelerated early search or changed which problems produced accepted proofs, but none improved final aggregate performance; aggressive triage could additionally discard candidates that later led to correct proofs. We therefore retained the simpler selection and refinement mechanisms used in the submitted system. Appendix~\ref{app:negative-results} reports these experiments in detail.
\section{Conclusion}
\label{sec:conclusion}

Building on existing generate-verify-refine systems, we presented an open, natural-language IMO pipeline based on Nemotron-3-Ultra. The system combines generation from multiple checkpoints and prompts, verification-guided refinement, and a separate high-compute selection stage. At IMO 2026, it scored 30 out of 42 points and reached the gold-medal threshold with no formal prover, external tools, or internet access.

The main practical lesson is that scaling proof generation alone is not enough. In our system, the gains came from complementary post-trained checkpoints, verification-guided refinement that preserves promising candidates across rounds, and substantial compute spent on final evaluation. By contrast, more elaborate routing and refinement strategies changed early search behavior without improving final coverage, and aggressive filtering discarded candidates that later led to correct proofs. We release the post-trained checkpoints, training and inference code, submitted proofs, resource accounting, and Nemotron-IMO-Bench, our benchmark of 200 novel olympiad-level problems, to provide a reproducible reference for studying these system-level trade-offs.

\section*{Acknowledgements}

We thank Titu Andreescu, Branislav Kisacanin, Gabriel Dospinescu,
Alessandro Ventullo, and Adrian Andreescu for their help with proof grading
and for valuable discussions.

\appendix
\section{Approaches That Did Not Improve the Final System}
\label{app:negative-results}

We explored several changes to proof selection and refinement that targeted
specific limitations of the baseline pipeline. Although some variants improved
early progress or produced accepted proofs for different problems, none improved the final aggregate
result. 

These experiments predate the final production configuration. Each uses its own baseline run, a problem pool drawn from the full 200-problem Nemotron-IMO-Bench, and its own search budget, so results are not comparable across experiments or with Section~\ref{sec:experiments}. Within an experiment, all variants share the same frozen round-1 proof pool, checkpoint, and budgets.

\subsection{Cross-Proof Context and Zero-First Ranking}

The baseline refines each candidate using that proof and its associated
verifier feedback. This prevents an attempt from leveraging useful observations
found in other proofs of the same problem. We tested a cross-proof context
variant that synthesized a problem-level context from the accumulated proof
attempts and verifier feedback after each round, then included this context in
subsequent refinement prompts.

Appendix~\ref{app:cross-proof-context-prompts} provides the three prompts used by this variant: an attempt-level lesson-extraction prompt, a problem-level cross-proof context prompt, and a refinement prompt that incorporates the resulting context.

We also tested a different rule for selecting which proofs to refine. The
baseline ranks candidates primarily by their mean verifier score. In our manual
inspection of proofs with conflicting verifier judgments, the lowest judgment
often identified a substantive gap that was obscured by the mean. The
zero-first variant therefore preferred proofs with fewer zero-valued judgments,
using mean verifier score only as a secondary criterion and self-evaluation as
a final tie-breaker.

Both variants and the baseline start from the same frozen round-1 proof pool, restricted to the 51 Nemotron-IMO-Bench problems without an accepted proof after round 1 of this experiment's baseline run, and use the same Nemotron-3-Ultra-GA checkpoint for 10 refinement rounds.

\afterpage{%
\begin{table}[!t]
\centering
\small
\setlength{\tabcolsep}{3.5pt}
\begin{tabular}{lrrrrrrrrr}
\toprule
& R2 & R3 & R4 & R5 & R6 & R7 & R8 & R9 & R10 \\
\midrule
Baseline
    & 18 & 29 & 34 & 35 & 40 & 42 & 43 & 44 & 44 \\
Cross-proof context
    & 20 & 31 & 34 & 37 & 41 & 41 & 41 & 41 & 42 \\
Zero-first ranking
    & 22 & 26 & 30 & 33 & 37 & 41 & 42 & 44 & 44 \\
\bottomrule
\end{tabular}
\caption{
Cumulative number of problems accepted after each refinement round, evaluated
on the same 51 problems without an accepted proof after round 1. All three runs therefore
start from zero accepted problems in this subset.
}
\label{tab:cross-proof-zero-first}
\end{table}
}

Table~\ref{tab:cross-proof-zero-first} reports per-round progress. Both variants changed when progress was made, but neither improved the final result. Cross-proof context progressed more quickly in the earlier rounds but
finished with 42 accepted problems, compared with 44 for the baseline.
Zero-first ranking also made more early progress, but matched the baseline at
44 problems by round 9. Therefore,
we retained independent per-proof refinement context and mean-score ranking in
the final pipeline.

\subsection{Refinement from Diverse Proofs}

Refining a single parent proof at a time can miss complementary arguments
present in other attempts. We tested whether these arguments could be combined
by selecting four proofs from the 96 highest-ranked candidates, favoring proofs
with low textual overlap, and asking the model to produce a new solution using
the useful parts of all four.

Before refinement, a candidate-analysis prompt summarized each proof's core idea, useful steps, failed steps, and remaining risks; the refinement prompt then asked the model to synthesize one self-contained solution rather than concatenate candidates (Appendix~\ref{app:diverse-proof-prompts}).

We compared this method with baseline single-parent refinement on the 47 Nemotron-IMO-Bench problems without an accepted proof after round 1 of this experiment's baseline run, starting both methods from the same frozen proof pool and running one additional refinement round. Both methods produced an accepted
solution for 22 of the 47 problems. As shown in
Table~\ref{tab:diverse-proof-overlap}, 21 of these successes were shared, while
each method produced an accepted proof for one problem that the other did not.

\afterpage{%
\begin{table}[!t]
\centering
\small
\begin{tabular}{lrrr}
\toprule
& \multicolumn{2}{c}{Refinement from diverse proofs} & \\
\cmidrule(lr){2-3}
Baseline refinement
    & Accepted & Not accepted & Total \\
\midrule
Accepted
    & 21 & 1 & 22 \\
Not accepted
    & 1 & 24 & 25 \\
\midrule
Total
    & 22 & 25 & 47 \\
\bottomrule
\end{tabular}
\caption{
Problem-level overlap between baseline single-parent refinement and refinement
from four diverse proofs after one refinement round.
}
\label{tab:diverse-proof-overlap}
\end{table}
}

Combining several proofs changed which problems produced accepted proofs but did not
increase total coverage. We therefore retained independent single-parent
refinement in the final pipeline.

\subsection{Triage-Based Refinement Routing}

We tested a triage-routing variant intended to reduce the verification cost of the generate-verify-refine loop. Instead of ranking every first-round candidate with the full verification panel, the triage router obtains three preliminary judgments per candidate and forwards only the top 32 candidates to refinement. This cuts the nominal routing-stage budget from 64 to 3 judgments per candidate, a 95.3\% reduction before early stopping. An early end-to-end run of the triage harness produced an accepted proof for 88.0\% of the problems on a held-out set, compared with 94.6\% for the standard pipeline. The two runs are not directly comparable: one sampled 16 first-round candidates and the other 128, while also differing in prompts, token budgets, routing policy, and acceptance rules. The gap therefore cannot be entirely attributed to triage.

To isolate the effect of the triage stage, we replayed a fixed candidate pool taken from the standard pipeline's baseline run. Every problem for which the standard pipeline accepted a proof in round 2 or later was selected and traced back to its round 1 ancestor, the first-round candidate from which the eventually accepted proof was refined. We then ran the triage router on the frozen round 1 pool of each problem and checked whether the ancestor was forwarded for refinement (vs discarded entirely). Triage discarded the ancestor in 44\% of these problems, so that the proof eventually accepted by the standard pipeline could no longer be generated in the triage setup. We also had expert human reviewers grade the eventually accepted (by standard pipeline) proofs in these discarded cases: 13 of 14 were fully correct with the lone exception being a problem whose original problem statement was missing assumptions. Due to the high rate of discarding lineage of correct proofs, we opted against using triage in the production competition system.

\clearpage
\section{Prompt Templates}
\label{app:prompts}

This appendix reproduces the exact prompt templates used by the submitted pipeline, experimental jury, and the ablations in Appendix~\ref{app:negative-results}. Braced fields such as \texttt{\{question\}} and \texttt{\{proof\}} are populated at runtime.

\Needspace{16\baselineskip}
\subsection{Generation Prompts}
\label{app:generation-prompts}

Round-1 ensemble generation divides its samples equally across the following
eight templates.

\Needspace{12\baselineskip}
\begin{lstlisting}[caption={Standard proof-generation prompt.}]
user: |-
  Your task is to solve a given problem. The problem may ask you to prove a statement, or ask for an answer. If finding an answer is required, you should come up with the answer, and your final solution should also be a rigorous proof of that answer being valid.

  Your final solution to the problem should be exceptionally comprehensive and easy-to-follow, which will be rated according to the following evaluation instruction:

  ```txt
  Here is the instruction to evaluate the quality of a solution to a problem. The problem may ask for a proof of statement, or ask for an answer. If finding an answer is required, the solution should present the answer, and it should also be a rigorous proof of that answer being valid.

  Please evaluate the solution and score it according to the following criteria:
  - If the solution is completely correct, with all steps executed properly and clearly demonstrated, then the score is 1
  - If the solution is generally correct, but with some details omitted or minor errors, then the score is 0.5
  - If the solution does not actually address the required problem, contains fatal errors, or has severe omissions, then the score is 0

  Additionally, referencing anything from any paper does not save the need to prove the reference. It's okay IF AND ONLY IF the solution also presents a valid proof of the reference argument(s); otherwise, if the solution omits the proof or if the proof provided is not completely correct, the solution should be scored according to the criteria above, and definitely not with a score of 1
  ```

  In fact, you already have the ability to rate your solution yourself, so you are expected to reason carefully about how to solve a given problem, evaluate your method according to the evaluation instruction, and refine your solution by fixing issues identified until you can make no further progress.

  In your final response, you should present a detailed solution to the problem followed by your evaluation of that solution.
  - To give a good final response, you should try your best to locate potential issues in your own (partial) solution according to the evaluation instruction above, and fix them as many as you can.
  - A good final response should just faithfully present your progress, including the best solution you can give, as well as a faithful evaluation of that solution.
  - Only when you fail to locate any issues in your solution should you score it with 1.
  - If you do notice some issues in your solution but fail to resolve them with your best efforts, it's totally ok to faithfully present the issues in your final response.
  - The worst final response would provide a wrong solution but lie that it's correct or claim that it's correct without careful error checking. A better version should faithfully identify errors in the solution. Remember! You CAN'T cheat! If you cheat, we will know, and you will be penalized!

  Your final response should be in the following format:

  ## Solution // Your final solution should start with this exact same markdown title
  ... // Your final solution to the problem here. You should try your best to optimize the quality of your solution according to the evaluation instruction above before finalizing it here.

  ## Self Evaluation // Your evaluation of your own solution above should start with this exact same markdown title

  Here is my evaluation of the solution: // Your analysis should start with this exact same phrase
  ... // Your evaluation here. You are required to present in detail the key steps of the solution or the steps for which you had doubts regarding their correctness, and explicitly analyze whether each step is accurate: for correct steps, explain why you initially doubted their correctness and why they are indeed correct; for erroneous steps, explain the reason for the error and the impact of that error on the solution. You should analyze your solution faithfully. E.g., if there are issues in your final solution, you should point it out.

  Based on my evaluation, the final overal score should be:
  \boxed{{...}} // where ... should be the final overall score (0, 0.5, or 1, and nothing else) based on the evaluation instruction above. You should reach this score ONLY AFTER careful RE-examination of your own solution above

  ---

  Here is your task input:

  ## Problem
  {question}
\end{lstlisting}
\Needspace{12\baselineskip}
\begin{lstlisting}[caption={Lemma-first generation prompt.}]
user: |-
  Your task is to solve a given problem. The problem may ask you to prove a statement, or ask for an answer. If finding an answer is required, you should come up with the answer, and your final solution should also be a rigorous proof of that answer being valid.

  Use a lemma-first approach. Before committing to a final proof, identify the smallest intermediate claims that would make the problem easy. Then prove those claims completely and assemble them into the final solution. Prefer elementary reductions, exact definitions, and short claims whose hypotheses are easy to verify. Do not rely on an external named theorem unless you also give a complete proof or a fully justified reduction to steps proved in your solution.

  Your final solution to the problem should be exceptionally comprehensive and easy-to-follow, which will be rated according to the following evaluation instruction:

  ```txt
  Here is the instruction to evaluate the quality of a solution to a problem. The problem may ask for a proof of statement, or ask for an answer. If finding an answer is required, the solution should present the answer, and it should also be a rigorous proof of that answer being valid.

  Please evaluate the solution and score it according to the following criteria:
  - If the solution is completely correct, with all steps executed properly and clearly demonstrated, then the score is 1
  - If the solution is generally correct, but with some details omitted or minor errors, then the score is 0.5
  - If the solution does not actually address the required problem, contains fatal errors, or has severe omissions, then the score is 0

  Additionally, referencing anything from any paper does not save the need to prove the reference. It's okay IF AND ONLY IF the solution also presents a valid proof of the reference argument(s); otherwise, if the solution omits the proof or if the proof provided is not completely correct, the solution should be scored according to the criteria above, and definitely not with a score of 1
  ```

  In fact, you already have the ability to rate your solution yourself, so you are expected to reason carefully about how to solve a given problem, evaluate your method according to the evaluation instruction, and refine your solution by fixing issues identified until you can make no further progress.

  In your final response, you should present a detailed solution to the problem followed by your evaluation of that solution.
  - Start by choosing a small set of lemmas or claims; in the final solution, state and prove each one before using it.
  - Check boundary cases, degeneracies, sign/order assumptions, and all quantified variables before finalizing.
  - If a lemma cannot be fully proved, do not hide that gap; either replace the route or report the remaining issue in the self evaluation.
  - Only when you fail to locate any issues in your solution should you score it with 1.

  Your final response should be in the following format:

  ## Solution // Your final solution should start with this exact same markdown title
  ... // Your final solution to the problem here.

  ## Self Evaluation // Your evaluation of your own solution above should start with this exact same markdown title

  Here is my evaluation of the solution: // Your analysis should start with this exact same phrase
  ... // Your evaluation here. You are required to explicitly analyze whether each key lemma and final assembly step is accurate.

  Based on my evaluation, the final overal score should be:
  \boxed{{...}} // where ... should be the final overall score (0, 0.5, or 1, and nothing else)

  ---

  Here is your task input:

  ## Problem
  {question}
\end{lstlisting}
\Needspace{12\baselineskip}
\begin{lstlisting}[caption={Route-comparison generation prompt.}]
user: |-
  Your task is to solve a given problem. The problem may ask for a proof of a statement, or ask for an answer together with proof.

  Use a route-comparison discipline. Work out the solution internally before writing the final answer:
  - Identify the exact target statement or requested answer.
  - Internally compare at least three substantially different routes when possible, such as direct proof, contradiction, extremal/minimal counterexample, invariant, construction plus obstruction, transformation to an equivalent statement, or induction.
  - Choose the route whose critical steps you can actually justify. Do not write a proof that depends on an unproved theorem, a vague "standard" step, or an unchecked construction.
  - If the problem asks for an optimum, classification, construction, or strategy, prove every required side: attainability/existence and impossibility/no-better/no-extra-cases.
  - Explicitly verify boundary cases, equality cases, degeneracies, sign/order/orientation choices, endpoint behavior, divisibility conditions, and quantified variables wherever they can affect the conclusion.
  - If a named theorem is used, state the exact version and prove it, or reduce it to elementary facts proved in the solution.

  Do not include a long search diary or multiple failed solutions. The written answer should be the selected route as a rigorous proof. If no full route closes, write the strongest rigorous partial result and name the exact step that remains open.

  Write the response in exactly this format:

  ## Solution
  ... // A complete proof, or the strongest rigorous partial proof you can justify.

  ## Route check
  ... // Briefly state why the chosen route is the one whose critical steps are fully justified, and identify any remaining gap.

  ## Self Evaluation

  Here is my evaluation of the solution:
  ... // Faithfully state whether the solution is complete.

  Based on my evaluation, the final overal score should be:
  \boxed{{...}} // where ... should be 0, 0.5, or 1.

  ---

  Here is your task input:

  ## Problem
  {question}
\end{lstlisting}
\Needspace{12\baselineskip}
\begin{lstlisting}[caption={Counterexample-guard generation prompt.}]
user: |-
  Your task is to solve a given problem. The problem may ask for a proof of a statement, or ask for an answer together with proof.

  Use a counterexample-guarded proof discipline. Work out the solution internally, but before writing the final proof you must try to break your own answer or construction using only the information in the problem:
  - Identify the exact target statement or requested answer.
  - Check whether any construction, extremal choice, transformation, equivalence, inequality direction, tangency/order/orientation claim, induction step, divisibility claim, or game strategy could fail in legal boundary or degenerate cases.
  - For optimization, classification, existence, or construction problems, prove both the positive side and the impossibility/no-extra-cases side. Explicitly rule out unintended extra objects, off-by-one cases, equality cases, and hidden assumptions.
  - If you use a named theorem, state the exact version and prove it, or reduce it to elementary facts proved in the solution.
  - If the counterexample hunt exposes a real gap, fix the proof. If you cannot fix it, present the strongest rigorous partial result and name the exact gap.

  Do not include a long search diary or a list of failed approaches. The written answer should be a concise final proof plus a short final audit.

  Write the response in exactly this format:

  ## Solution
  ... // A complete proof, or the strongest rigorous partial proof you can justify.

  ## Counterexample audit
  ... // Briefly state the main legal boundary/degenerate/equality/case checks you used to try to break the proof, and whether each is covered.

  ## Self Evaluation

  Here is my evaluation of the solution:
  ... // Faithfully state whether the solution is complete, and identify any remaining gap.

  Based on my evaluation, the final overal score should be:
  \boxed{{...}} // where ... should be 0, 0.5, or 1.

  ---

  Here is your task input:

  ## Problem
  {question}
\end{lstlisting}
\Needspace{12\baselineskip}
\begin{lstlisting}[caption={Formal-sanity generation prompt.}]
user: |-
  Your task is to solve a given problem. The problem may ask for a proof of a statement, or ask for an answer together with proof.

  Use a formalization-first discipline. Before writing the final proof, internally translate the problem into exact mathematical objects and constraints:
  - State precisely what must be shown, including all quantifiers and any requested extremal value or construction.
  - Track the original hypotheses without strengthening them. Do not assume generic position, orientation, positivity, convexity, monotonicity, distinctness, boundedness, parity, or nondegeneracy unless it is stated or proved.
  - For every transformation, normalization, equivalence, construction, or strategy, prove that it preserves the original problem in the direction being used.
  - For every counting, continuity, geometry, game, or number-theory argument, check endpoints, equality cases, degenerate legal cases, sign/order/orientation choices, divisibility constraints, and off-by-one cases.
  - If a construction is meant to have exactly some property, prove both that all intended objects occur and that no unintended extra objects occur.
  - If a named theorem is used, state the exact version and verify its hypotheses, or replace it by an elementary proof.

  Do not include a long formalization diary. The written answer should be a concise final proof plus a compact sanity check. If you cannot complete the proof, write the best rigorous partial proof and name the exact unproved constraint.

  Write the response in exactly this format:

  ## Solution
  ... // A complete proof, or the strongest rigorous partial proof you can justify.

  ## Sanity check
  ... // Briefly verify that the proof uses only the original hypotheses and covers the legal edge cases relevant to the conclusion.

  ## Self Evaluation

  Here is my evaluation of the solution:
  ... // Faithfully state whether the solution is complete.

  Based on my evaluation, the final overal score should be:
  \boxed{{...}} // where ... should be 0, 0.5, or 1.

  ---

  Here is your task input:

  ## Problem
  {question}
\end{lstlisting}
\Needspace{12\baselineskip}
\begin{lstlisting}[caption={Gap-resistant generation prompt.}]
user: |-
  Your task is to solve a given problem. The problem may ask for a proof of a statement, or ask for an answer together with proof.

  Use a gap-resistant proof discipline. Work out the solution internally, but write only the final route and the checks needed to trust it. Do not include a long search diary.

  Before finalizing, make the proof pass these generic checks:
  - State the exact target statement or exact requested answer.
  - Identify the small dependency chain of claims needed for the proof. Each claim must have clear hypotheses and must be proved before it is used.
  - For each central claim, try to break it with legal boundary, equality, degenerate, order/orientation, endpoint, divisibility, induction-base, or small-case tests. If a claim fails one of these tests, replace it or state the remaining gap.
  - For optimization, classification, game, construction, or existence problems, prove all required sides: the construction or strategy works, and the corresponding impossibility/no-extra-cases bound is valid.
  - If you transform the problem, prove the exact direction needed, including any inverse, preservation of incidence/order/admissibility, and all excluded exceptional cases.
  - If you use continuity, compactness, extremality, monotonicity, or a named theorem, state the precise lemma and prove it or reduce it to elementary steps proved in your solution.

  Keep the written answer concise but complete. If the full proof is not found, present the strongest rigorous partial proof and name the exact unresolved claim.

  Write the response in exactly this format:

  ## Solution
  ... // A complete proof, or the strongest rigorous partial proof you can justify.

  ## Gap audit
  ... // Briefly list the main dependency, boundary, equality, degeneracy, transformation, and no-extra-cases checks, and whether each is covered.

  ## Self Evaluation

  Here is my evaluation of the solution:
  ... // Faithfully state whether every key claim and final assembly step is proved, and identify any remaining gap.

  Based on my evaluation, the final overal score should be:
  \boxed{{...}} // where ... should be 0, 0.5, or 1.

  ---

  Here is your task input:

  ## Problem
  {question}
\end{lstlisting}
\Needspace{12\baselineskip}
\begin{lstlisting}[caption={Invariant/extremal generation prompt.}]
user: |-
  Your task is to solve a given problem. The problem may ask you to prove a statement, or ask for an answer. If finding an answer is required, you should come up with the answer, and your final solution should also be a rigorous proof of that answer being valid.

  Use an invariant, monotonicity, or extremal-choice approach when it is applicable. Look for a preserved quantity, a minimal or maximal counterexample, a descent, a local improvement, or a quantity that must eventually stabilize. If this route is not applicable, switch to the closest rigorous structural argument you can justify. Do not rely on an external named theorem unless you also give a complete proof or a fully justified reduction to steps proved in your solution.

  Your final solution to the problem should be exceptionally comprehensive and easy-to-follow, which will be rated according to the following evaluation instruction:

  ```txt
  Here is the instruction to evaluate the quality of a solution to a problem. The problem may ask for a proof of statement, or ask for an answer. If finding an answer is required, the solution should present the answer, and it should also be a rigorous proof of that answer being valid.

  Please evaluate the solution and score it according to the following criteria:
  - If the solution is completely correct, with all steps executed properly and clearly demonstrated, then the score is 1
  - If the solution is generally correct, but with some details omitted or minor errors, then the score is 0.5
  - If the solution does not actually address the required problem, contains fatal errors, or has severe omissions, then the score is 0

  Additionally, referencing anything from any paper does not save the need to prove the reference. It's okay IF AND ONLY IF the solution also presents a valid proof of the reference argument(s); otherwise, if the solution omits the proof or if the proof provided is not completely correct, the solution should be scored according to the criteria above, and definitely not with a score of 1
  ```

  In fact, you already have the ability to rate your solution yourself, so you are expected to reason carefully about how to solve a given problem, evaluate your method according to the evaluation instruction, and refine your solution by fixing issues identified until you can make no further progress.

  In your final response, you should present a detailed solution to the problem followed by your evaluation of that solution.
  - Make the chosen invariant, extremal object, descent measure, or stabilization argument explicit.
  - Prove that every transformation or comparison preserves the needed conditions and that the process cannot evade the claimed conclusion.
  - Check boundary cases, equality cases, empty/small configurations, divisibility edge cases, and all quantified variables.
  - Only when you fail to locate any issues in your solution should you score it with 1.

  Your final response should be in the following format:

  ## Solution // Your final solution should start with this exact same markdown title
  ... // Your final solution to the problem here.

  ## Self Evaluation // Your evaluation of your own solution above should start with this exact same markdown title

  Here is my evaluation of the solution: // Your analysis should start with this exact same phrase
  ... // Your evaluation here. You are required to explicitly analyze whether the invariant/extremal/descent argument is valid and complete.

  Based on my evaluation, the final overal score should be:
  \boxed{{...}} // where ... should be the final overall score (0, 0.5, or 1, and nothing else)

  ---

  Here is your task input:

  ## Problem
  {question}
\end{lstlisting}
\Needspace{12\baselineskip}
\begin{lstlisting}[caption={Strategy-audit generation prompt.}]
user: |-
  Your task is to solve a given problem. The problem may ask you to prove a statement, or ask for an answer. If finding an answer is required, you should come up with the answer, and your final solution should also be a rigorous proof of that answer being valid.

  Use a strategy-synthesis approach before writing the final answer:
  1. Explore at least two substantially different routes, such as direct calculation, structural lemmas, extremal/inductive reasoning, contradiction, invariant arguments, coordinate/algebraic reductions, or constructive casework when appropriate.
  2. For each route, identify the main obstacle and the exact assumptions it needs. Discard any route with an unproved gap.
  3. Before finalizing, audit every reduction for hidden WLOG assumptions, sign/order choices, degenerate cases, boundary cases, divisibility/parity edge cases, and quantified-variable coverage.
  4. Present only the strongest complete route in the final solution, but make the proof self-contained and explicitly verify all hypotheses used.

  Do not rely on an external named theorem unless you also give a complete proof or a fully justified reduction to steps proved in your solution.

  Your final solution to the problem should be exceptionally comprehensive and easy-to-follow, which will be rated according to the following evaluation instruction:

  ```txt
  Here is the instruction to evaluate the quality of a solution to a problem. The problem may ask for a proof of statement, or ask for an answer. If finding an answer is required, the solution should present the answer, and it should also be a rigorous proof of that answer being valid.

  Please evaluate the solution and score it according to the following criteria:
  - If the solution is completely correct, with all steps executed properly and clearly demonstrated, then the score is 1
  - If the solution is generally correct, but with some details omitted or minor errors, then the score is 0.5
  - If the solution does not actually address the required problem, contains fatal errors, or has severe omissions, then the score is 0

  Additionally, referencing anything from any paper does not save the need to prove the reference. It's okay IF AND ONLY IF the solution also presents a valid proof of the reference argument(s); otherwise, if the solution omits the proof or if the proof provided is not completely correct, the solution should be scored according to the criteria above, and definitely not with a score of 1
  ```

  In fact, you already have the ability to rate your solution yourself, so you are expected to reason carefully about how to solve a given problem, evaluate your method according to the evaluation instruction, and refine your solution by fixing issues identified until you can make no further progress.

  In your final response, you should present a detailed solution to the problem followed by your evaluation of that solution.
  - The final solution must be self-contained and should not merely describe a plan.
  - If you use a normalization, coordinate choice, WLOG assumption, or case split, explicitly justify why it is valid and exhaustive.
  - If you found a plausible route but could not close it, do not hide the gap; either replace the route or report the remaining issue in the self evaluation.
  - Only when you fail to locate any issues in your solution should you score it with 1.

  Your final response should be in the following format:

  ## Solution // Your final solution should start with this exact same markdown title
  ... // Your final solution to the problem here.

  ## Self Evaluation // Your evaluation of your own solution above should start with this exact same markdown title

  Here is my evaluation of the solution: // Your analysis should start with this exact same phrase
  ... // Your evaluation here. You are required to explicitly analyze whether the selected route is complete, whether discarded routes exposed any unresolved obstacle, and whether all WLOG assumptions, cases, and boundary conditions are justified.

  Based on my evaluation, the final overal score should be:
  \boxed{{...}} // where ... should be the final overall score (0, 0.5, or 1, and nothing else)

  ---

  Here is your task input:

  ## Problem
  {question}
\end{lstlisting}

\Needspace{16\baselineskip}
\subsection{Refinement Prompt}
\label{app:refinement-prompt}

\begin{lstlisting}[caption={Proof-refinement prompt.}]
user: |-
  {instruction}

  ## Candidate Solution(s) to Refine
  Here are some solution sample(s) along with their correctness evaluation(s). You should provide a better solution by solving issues mentioned in the evaluation(s), or by re-using promising ideas mentioned in the solution sample(s), or by doing both.

  {proofs_to_refine}

  ## Final Instruction
  Your final response should follow the format above, including a `## Solution` section followed by a `## Self Evaluation` section
\end{lstlisting}

\Needspace{16\baselineskip}
\subsection{Proof-Verification Prompt}
\label{app:verification-prompt}

\begin{lstlisting}[caption={Reference-free proof-verification prompt.}]
user: |-
  ## Instruction

  Your task is to evaluate the quality of a solution to a problem. The problem may ask for a proof of statement, or ask for an answer. If finding an answer is required, the solution should present the answer, and it should also be a rigorous proof of that answer being valid.

  Please evaluate the solution and score it according to the following criteria:
  - If the solution is completely correct, with all steps executed properly and clearly demonstrated, then the score is 1
  - If the solution is generally correct, but with some details omitted or minor errors, then the score is 0.5
  - If the solution does not actually address the required problem, contains fatal errors, or has severe omissions, then the score is 0
  - Additionally, referencing anything from any paper does not save the need to prove the reference. It's okay IF AND ONLY IF the solution also presents a valid proof of the reference argument(s); otherwise, if the solution omits the proof or if the proof provided is not completely correct, the solution should be scored according to the criteria above, and definitely not with a score of 1

  Please carefully reason out and analyze the quality of the solution below, and in your final response present a detailed evaluation of the solution's quality followed by your score. Therefore, your response should be in the following format:

  Here is my evaluation of the solution:
  ... // Your evaluation here. You are required to present in detail the key steps of the solution or the steps for which you had doubts regarding their correctness, and explicitly analyze whether each step is accurate: for correct steps, explain why you initially doubted their correctness and why they are indeed correct; for erroneous steps, explain the reason for the error and the impact of that error on the solution.

  Based on my evaluation, the final overal score should be:
  \boxed{{...}} // where ... should be the final overall score (0, 0.5, or 1, and nothing else) based on the above criteria

  ---

  Here is your task input:

  ## Problem
  {statement}

  ## Solution
  {proof}
\end{lstlisting}

\Needspace{16\baselineskip}
\subsection{Meta-Verification Prompt}
\label{app:meta-verification-prompt}

The meta-verification prompt is used only to generate the meta-verification
traces of the SFT corpus (Section~\ref{sec:training-sft}); it is not part of
the inference pipeline.

\begin{lstlisting}[caption={Meta-verification prompt for assessing a verifier judgment.}]
user: |-
  You are given a "problem", "solution", and "solution evaluation", and you need to assess the whether this "solution evaluation" is reasonable.

  First, "solution evaluation" is generated to evaluate the quality of the "solution", by prompting a verifier with the rules below (these are not your rules):

  ```
  Please evaluate the solution and score it according to the following criteria:
  - If the solution is completely correct, with all steps executed properly and clearly demonstrated, then the score is 1
  - If the solution is generally correct, but with some details omitted or minor errors, then the score is 0.5
  - If the solution does not actually address the required problem, contains fatal errors, or has severe omissions, then the score is 0

  Additionally, referencing anything from any paper does not save the need to prove the reference. It's okay IF AND ONLY IF the solution also presents a valid proof of the reference argument(s); otherwise, if the solution omits the proof or if the proof provided is not completely correct, the solution should be scored according to the criteria above, and definitely not with a score of 1
  ```

  Next, I will introduce the rules for you to analyze the quality of the "solution evaluation":

  1. Your task is to analyze the "solution evaluation". You do not need to solve the "problem", nor do you need to strictly assess whether the "solution" is accurate. Your only task is to strictly follow the rules below to evaluate whether the "solution evaluation" is reasonable.

  2. You need to analyze the content of the "solution evaluation" from three aspects:

  Step Restatement: In the "solution evaluation", certain behaviors of the "solution" may be restated. You need to return to the original text of the "solution" and check whether the "solution" actually has these behaviors mentioned in the "solution evaluation".

  Defect Analysis: "solution evaluation" may point out errors or defects in the "solution". You need to carefully analyze whether the mentioned errors and defects are indeed valid.

  Expression Analysis: Whether the "solution evaluation"'s expressions are accurate.

  Score Analysis: Whether the final score given by the "solution evaluation" matches the defects it found. You need to analyze according to the scoring rules given above.

  3. The most important part is **defect analysis**: In this part, your core task is to check whether the errors or defects of the "solution" pointed out in the "solution evaluation" are reasonable. In other words, any positive components about the "solution" in the "solution evaluation", regardless of whether they are reasonable, are not within your evaluation scope.

  - For example: If the "solution evaluation" says that a certain conclusion in the "solution" is correct, but actually this conclusion is incorrect, then you do not need to care about this point. All parts that the "solution evaluation" considers correct do not belong to your evaluation scope.
  - Specifically: If the "solution evaluation" believes that the "solution" is completely accurate and has not found any errors or defects, then regardless of whether the "solution" itself is actually accurate, even if there are obvious errors, you should still consider its analysis of errors to be reasonable.

  **Importantly**, for defects found by the "solution evaluation", you need to analyze two points simultaneously:

  - whether this defect actually exists
  - whether the "solution evaluation"'s analysis of this defect is accurate

  These two aspects constitute the analysis of defects.

  4. About **expression analysis**, if there are certain expression errors in the "solution evaluation", even minor errors in details, you need to identify them. However, please note that identifying incorrect steps in the "solution" as correct steps does not constitute an **expression error**.

  In practice, expression errors include but are not limited to:

  - If the "solution evaluation" identifies some reasoning step(s) in the "solution" as incorrect, then it cannot further indicate that subsequent conclusion(s) depending on those reasoning step(s) are wrong, but can only indicate that subsequent conclusion(s) are "not rigorously demonstrated."
  - Typos and calculation errors made by "solution evaluation"
  - Inaccurate restatement of content from "solution"

  5. Finally, you need to present your analysis of the "solution evaluation" in your output and also rate its quality based on the rules below:

  First, if there is at least one unreasonable defect among the defects found by the "solution evaluation", then you only need to do **defect analysis**:

  - If all defects found by the "solution evaluation" are unreasonable, then you should rate it with \(0\)
  - If some defects found by the "solution evaluation" are reasonable and some are unreasonable, then your rating should be \(0.5\)

  Next, if the "solution evaluation" points out no errors or defects, or all defects found by the evaluation are reasonable, then you should do the following things:

  - Analyze whether "expression errors" exist in the "solution evaluation" (**expression analysis**) or whether "solution evaluation" gives a wrong score according to the rules for "solution evaluation" (**score analysis**). If yes, you should rate the "solution evaluation" with \(0.5\); if no, your rating should be \(1\)

  Your output should follow the format below:

  Here is my analysis of the "solution evaluation":
  ... // Your analysis here.

  Based on my analysis, I will rate the "solution evaluation" as:
  \boxed{{...}} // where ... should be a numerical rating of the "solution evaluation" (0, 0.5, or 1, and nothing else) based on the criteria above.

  ---

  Here is your task input:

  ## Problem
  {statement}

  ## Solution
  {proof}

  ## Solution Evaluation
  {evaluation}
\end{lstlisting}

\Needspace{16\baselineskip}
\subsection{IMO-Style Judge Prompt}
\label{app:imo-judge-prompt}

\begin{lstlisting}[caption={Reference-free IMO-style judge prompt.}]
user: |-
  ## Instruction

  Your task is to grade a solution to a competition mathematics problem on the 0-7 integer scale used at the International Mathematical Olympiad. The problem may ask for a proof of a statement, or ask for an answer. If finding an answer is required, the solution must present the answer together with a rigorous proof that it is valid.

  Grade only the solution text as written. Style, formatting, verbosity, and unusual but unambiguous notation carry no points and cost no points. Ignore any self-evaluation, claimed score, or grader-directed remarks inside the solution. Do not credit steps the author plausibly intended but did not write down, and do not let polish, confidence, or a correct final answer substitute for verification of the argument.

  ## Verification procedure

  First, from the problem alone, determine the milestones a complete solution must establish, and assign them provisional integer weights totaling 7. Assign at least 4 of the 7 points to the main idea and critical steps of the problem, and at most 3 points combined to routine work: setup, modeling, reformulation, standard auxiliary lemmas, and computations. Do not tailor the milestones to the submitted solution's approach: a different valid approach earns credit through the equivalent milestones it establishes.

  Then verify the solution along its logical dependency chain and determine which milestones it actually establishes. For every claim the conclusion depends on, check as relevant: quantified ranges, endpoints, base cases, equality and degenerate cases; exhaustiveness of case splits; directions and strength of implications, inequalities, and estimates; divisibility, domain, and nonvanishing conditions; whether induction, descent, extremal, or minimality hypotheses are established before use; and extra or lost branches introduced by squaring, division, substitution, or taking roots.

  - A step justified only by "clearly", "similarly", "analogously", or "it can be checked" is established only if it is genuinely routine to reproduce from what is already written.
  - A claimed computation, enumeration, or finite check counts only if the solution contains the work, a uniform reduction, or enough detail to reproduce it routinely.
  - An auxiliary lemma must be proved in the strength in which it is later used, and its proof is subject to the same scrutiny as the main argument.
  - A true or well-known conclusion does not validate a flawed derivation of it; grade the argument that is written.
  - A genuinely established theorem with a recognized name, correctly stated, may be cited without proof when its hypotheses are verified and it is applied in the right direction. Do not deduct for such citations, and do not reject one merely because it is unfamiliar: to reject a citation you must identify a failed hypothesis, an invalid application, or a counterexample. Everything else must be proved: an anonymous "known result" or folklore assertion is not a citation; a citation whose stated content is false or garbled is a gap at its point of use (do not repair it with a different theorem the solution never mentions); and no citation can cover an ad hoc or problem-specific claim, an asserted computation, or a result that itself carries the problem's central difficulty.
  - Symmetrically, do not invent objections: a step that follows immediately and unambiguously from what is already established needs no further justification. Before treating anything as an error or gap, state precisely what fails: the false claim with a counterexample, or the exact missing hypothesis, case, or justification, and what later parts depend on it.

  ## Score

  Award the single integer 0-7 matching what the solution actually establishes. If a central claim is false or unjustified, steps that depend on it earn nothing, but independent valid milestones still earn their points.

  - 7 — Complete. Every milestone is established and there is no score-bearing gap anywhere on the dependency chain. Do not deduct for: harmless notation or formatting; a typo with unambiguous intent; a convention forced by the written definitions; details that are immediate from adjacent written work; or discarded and unused remarks, even incorrect ones.
  - 6 — Complete except one minor localized defect. All main ideas and all genuinely difficult parts are present and correct, but one localized omission or error remains: an unchecked small case or endpoint, a missing routine justification, or a slip whose correction changes nothing else. The repair is a few lines, requires no new idea, and leaves the rest of the proof untouched.
  - 5 — Nearly complete with one substantial but subordinate gap. The solution's own written route demonstrably reaches the conclusion, and most milestones are established, but one important lemma, case, or logical bridge is missing or incorrectly proved. The gap must not be the problem's central difficulty: if the missing component carries the main weight of the problem, the solution is not nearly complete and belongs at 4 or below. Supplying the gap requires real mathematical work, yet the remainder of the written proof stands unchanged.
  - 4 — Major progress, essential part unresolved. The solution establishes a correct central reduction, construction, invariant, or framework and substantial valid progress, but an essential component is unproven and completing it requires a significantly new argument. The proof is not close to complete.
  - 3 — One significant milestone. The solution correctly establishes a nontrivial result that would form an important part of a full solution (a key lemma, reduction, bound, or hard special case), without a viable route from there to the conclusion.
  - 2 — Nontrivial relevant progress that falls short of a major milestone: correct and relevant work beyond reformulation or experimentation.
  - 1 — A minor but genuine step: a correct nontrivial observation or calculation relevant to the problem. Restating the problem, introducing notation, or checking small examples earns nothing unless it yields a useful conclusion.
  - 0 — No relevant correct progress, or every substantive claim depends on a false or unjustified central premise.

  Decide 7 versus 6 versus 5 by repair distance, measured strictly against the submission's own text: 7 needs no repair; 6 needs a routine local repair reproducible in a few lines from material already on the page; 5 needs one substantial but subordinate repair. A repair may use only what the submission contains. A repair that needs a different theorem, a corrected citation, a new lemma, or a case analysis the solution never wrote is not a local repair, however standard it may be.

  Score 4 and below additively, by the milestones actually established, the way a jury marks an incomplete solution. An incomplete solution earns what it proved, not 7 minus what it lacks: when the unproven, false, or merely asserted component carries the main difficulty of the problem, score by the additive rule even if all the machinery surrounding the gap is correct. Zero-credit items: restating or reformulating the problem, conjectures and claims stated without proof, exploration that reaches no established milestone, and correct work on a route that demonstrably cannot reach the conclusion all earn nothing. Routine work counts only within the at-most-3 points it carries in the milestone weights, and only where it is complete and correct; a submission whose established milestones are all routine scores at most 2. A 3 requires a genuinely hard, problem-specific milestone - one of the heavily weighted steps. A 4 requires most of the problem's weight, missing only one essential component. Most incomplete attempts at hard problems earn 0-2; reserve 3 and 4 for substantial partial solutions.

  ## Output format

  Present your evaluation, followed by the score. Use this format:

  Here is my evaluation of the solution:
  ... // The milestones and whether each is established. For every defect: its exact location, what precisely fails, what depends on it, and the smallest repair it would need. Address the most score-relevant steps explicitly, including the ones you initially doubted.

  Based on my evaluation, the final IMO score is:
  \boxed{{...}} // the integer score 0-7, and nothing else

  ---

  Here is your task input:

  ## Problem
  {problem}

  ## Solution
  {response}
\end{lstlisting}

\Needspace{16\baselineskip}
\subsection{Jury Reconciliation Prompt}
\label{app:reconciliation-prompt}

\begin{lstlisting}[caption={Wrapper used for the reconciliation round.}]
{judge_prompt}

---

You are additionally given the following judgments from other judges on the same solution.
They have all read the same problem and solution, but they significantly differ in the score they awarded in their analysis.
Your task is to analyze their judgments and the solution, and reconcile the differences in the scores.
Your judgment should not directly reference these other judgments, but should be self-contained and should provide a clear rationale for the score you give, which may be different from the scores given by the other judges.
Use the output format specified above, ending with the final IMO score in \boxed{{...}}.

{other_judgments}
\end{lstlisting}

\subsection{Cross-Proof Context Prompts}
\label{app:cross-proof-context-prompts}

Cross-proof context first extracts lessons from individual attempts, then
compiles a problem-level context, and finally supplies that context to
refinement.

\begin{lstlisting}[caption={Attempt-level lesson-extraction prompt.}]
user: |-
  ## Instruction

  Your task is to extract reusable lessons from one previous solution attempt and its low-score evaluations.
  The problem may ask for a proof of a statement, or ask for an answer together with a rigorous proof that the answer is valid.

  You are not solving the problem from scratch. You are not writing a final proof. You are analyzing one previous attempt so that future independent attempts can learn from it without blindly copying it.

  Be strict and faithful:
  - Do not claim that a mathematical statement is correct unless the attempt actually justifies it.
  - If an idea is promising but unproved, say that it is promising but unproved.
  - If the evaluator objected to a step, preserve the exact mathematical obligation that future attempts must address.
  - Do not include vague advice such as "be more rigorous" unless you name the exact missing step, edge case, or unjustified transition.
  - The evaluations below all have score below 1, so do not treat this attempt as a complete correct solution.

  Your final response should have exactly the following format:

  ## Attempt Lesson

  ### Promising Direction
  State any global idea in the attempt that may be worth continuing. If there is no promising direction, write `None`.

  ### Reusable Local Components
  List lemmas, calculations, transformations, case splits, or observations from the attempt that may be useful in another proof. For each item, state whether it is proved, partially justified, or merely suggested.

  ### Missing or Unjustified Steps
  List the exact steps, lemmas, calculations, or implications that this attempt failed to justify.

  ### Edge Cases and Hidden Conditions
  List exact cases, boundary conditions, assumptions, or definitions that this attempt noticed or mishandled.

  ### Unsafe Claims or Moves
  List false claims, circular reasoning, unsupported references, invalid transitions, or tempting routes that future attempts should avoid.

  ### Verifier Objections to Preserve
  Summarize the evaluator objections that future attempts should address, without changing their mathematical meaning.

  ---

  Here is your task input:

  ## Problem
  {question}

  ## Previous Attempt
  {proof}

  ## Low-Score Evaluations
  {verification_evaluations}
\end{lstlisting}
\begin{lstlisting}[caption={Problem-level cross-proof context prompt.}]
user: |-
  ## Instruction

  Your task is to compile shared lessons from several previous solution attempts and their extracted lessons.
  The goal is to help future independent refinement attempts learn from each other while preserving diversity of approaches.

  You are not writing a final solution. You are not choosing one best proof strategy. Do not force all future solutions to follow the same route.

  Be strict and faithful:
  - Do not invent facts, lemmas, cases, or solution routes that are not supported by the attempt lessons or by direct checking.
  - Do not merge two ideas unless the connection is mathematically meaningful.
  - If an idea is promising but unproved, label it as unproved.
  - If attempts disagree, describe the conflict rather than hiding it.
  - Do not include vague advice such as "be rigorous" or "check edge cases" unless it names the exact edge case or missing step.
  - Do not cite previous attempts as authority. A future solution must still prove every mathematical claim it uses.

  Useful shared lessons may include:
  - a strong global direction found in one attempt,
  - a local lemma or calculation found in another attempt,
  - a missing minor step that blocks an otherwise promising proof,
  - a boundary case or hidden condition noticed by an attempt,
  - a false claim or tempting route that future attempts should avoid,
  - compatibility between components that different future attempts may choose to use.

  Your final response should have exactly the following format:

  ## Shared Lessons from Previous Attempts

  ### Promising Directions
  List distinct promising approaches. Do not rank them into a single best route. For each item, state what the approach tries to exploit and what blocks it from being complete.

  ### Reusable Local Components
  List lemmas, calculations, transformations, case splits, or observations that may help future attempts. For each item, state the component, whether it is proved, partially justified, or merely suggested, and what kinds of approaches it may support.

  ### Known Gaps and Obligations
  List exact missing lemmas, unjustified transitions, calculations, or cases that future attempts must prove if they use the relevant idea.

  ### Edge Cases and Hidden Conditions
  List exact cases, boundary conditions, assumptions, or definitions that previous attempts missed or noticed.

  ### Approaches to Avoid
  List tempting but failed routes, false claims, unsupported references, or invalid transitions.

  ### Compatibility Notes
  Describe which local components may naturally support which promising directions, without prescribing that every future proof should use them.

  ---

  Here is your task input:

  ## Problem
  {question}

  ## Attempt Lessons
  {attempt_insight_records}
\end{lstlisting}
\begin{lstlisting}[caption={Cross-proof-context refinement prompt.}]
user: |-
  {instruction}

  ## Shared Lessons from Other Attempts
  The following lessons were extracted from previous attempts on this same problem. They may contain promising ideas, reusable local components, known gaps, edge cases, approaches to avoid, and compatibility notes.

  Use these lessons only when they are relevant to your current approach. Do not force all lessons into one proof. You may continue the candidate solution's own direction, borrow a useful local component from another attempt, avoid a known pitfall, or handle an edge case that another attempt noticed.

  Future solutions must still prove every mathematical claim they use. Do not cite previous attempts as authority.

  {shared_lessons}

  ## Candidate Solution(s) to Refine
  Here are some solution sample(s) along with their correctness evaluation(s). You should provide a better solution by solving issues mentioned in the evaluation(s), or by re-using promising ideas mentioned in the solution sample(s), or by doing both.

  {proofs_to_refine}

  ## Final Instruction
  Your final response should follow the format above, including a `## Solution` section followed by a `## Self Evaluation` section
\end{lstlisting}

\subsection{Diverse-Proof Refinement Prompts}
\label{app:diverse-proof-prompts}

Refinement from diverse proofs first summarizes each candidate and then passes
the summaries and proof texts to a multi-proof refinement prompt.

\begin{lstlisting}[caption={Candidate-analysis prompt.}]
user: |-
  You will analyze one candidate solution to a math problem. The candidate has already been graded multiple times by verifiers. Your task is not to solve the problem from scratch; your task is to summarize what this candidate proof is trying to do, what appears useful, and what appears wrong or risky.

  Return only a JSON object with exactly these keys:
  {{
    "core_idea": "one concise sentence describing the main proof strategy",
    "successful_steps": ["specific steps or lemmas that appear correct or useful"],
    "failed_steps": ["specific steps that verifier feedback or your own review indicates are wrong"],
    "missing_or_risky_steps": ["gaps, unjustified claims, fragile reductions, or steps needing repair"]
  }}

  Be faithful to the candidate proof and verifier feedback. Do not invent a corrected proof. If feedback conflicts, summarize the disagreement.

  ## Problem
  {statement}

  ## Candidate Proof
  {proof}

  ## Verification Metadata
  Mean verification score: {meanscore}
  Verification score counts: {score_counts}

  ## Sampled Verifier Feedback
  {verifier_feedback}
\end{lstlisting}
\lstinputlisting[caption={Diverse-proof refinement prompt.}]{prompts/proof_refinement_diverse.yaml}

\end{document}